\pdfoutput=1
\documentclass[11pt]{article}

\usepackage[preprint]{acl}
\usepackage{nicematrix}
\usepackage{times}
\usepackage{latexsym}
\usepackage{subfigure}
\usepackage{amsmath}
\usepackage{amsfonts}
\usepackage{enumitem}
\usepackage{multirow}
\usepackage{colortbl} 
\usepackage{xcolor}   
\usepackage{booktabs}
\usepackage{adjustbox}
\usepackage{makecell}
\usepackage{hyperref}
\usepackage{tabularx}
\usepackage{booktabs}
\usepackage{array}
\usepackage[linesnumbered,ruled,vlined]{algorithm2e}

\usepackage{booktabs}
\newcommand{\Thickline}{\noalign{\hrule height 1.2pt}}
\usepackage[T1]{fontenc}

\usepackage[utf8]{inputenc}

\usepackage{microtype}

\usepackage{inconsolata}

\usepackage{graphicx}

\newcommand\ourmethod{\texttt{SARChat}\xspace}

\title{SARChat-Bench-2M: A Multi-Task Vision-Language Benchmark for SAR Image Interpretation}

\author{
 Zhiming~Ma$^{1,2\dag}$, Xiayang~Xiao$^{1,2\dag*}$, Sihao~Dong$^{3\dag}$, Peidong~Wang$^4$, HaiPeng~Wang$^{1*}$, Qingyun~Pan$^{5}$  \hspace{0.4em} \\
 $^1$The Key Laboratory for Information Science of Electromagnetic Waves (Ministry of Education), \\ School of Information Science and Technology, Fudan University, Shanghai, China \\
 $^2$China Mobile Internet Company Ltd., Guangzhou, China \\
 $^3$The School of Automation and Electrical Engineering, \\ Inner Mongolia University of Science and Technology, Baotou, China \\
 $^4$School of Computer Science and Engineering, Northeastern University, Shenyang, China \\
 $^5$China Mobile Group Guangdong Co., Ltd. Guangzhou Branch, Guangzhou, China \\
}

\begin{document}
\maketitle
\begin{abstract}

As a powerful all-weather Earth observation tool, synthetic aperture radar (SAR) remote sensing enables critical military reconnaissance, maritime surveillance, and infrastructure monitoring. Although Vision language models (VLMs) have made remarkable progress in natural language processing and image understanding, their applications remain limited in professional domains due to insufficient domain expertise. This paper innovatively proposes the first large-scale multimodal dialogue dataset for SAR images, named SARChat-2M, which contains approximately 2 million high-quality image-text pairs, encompasses diverse scenarios with detailed target annotations. This dataset not only supports several key tasks such as visual understanding and object detection tasks, but also serves as the first visual-language benchmark in the SAR domain. Through this work, we enable and evaluate VLMs' capabilities in SAR image interpretation, providing a paradigmatic framework for constructing multimodal datasets across various remote sensing vertical domains. Through experiments on 16 mainstream VLMs, the effectiveness of the dataset has been fully verified. The project will be released at 
\url{https://github.com/JimmyMa99/SARChat}.


\end{abstract}

\def\thefootnote{\dag}\footnotetext{These authors contributed equally to this work.}\def\thefootnote{\arabic{footnote}}
\def\thefootnote{*}\footnotetext{Corresponding author. Xiayang~Xiao and Haipeng~Wang}
\def\thefootnote{*}\footnotetext{Contact email: mazhiming312@outlook.com or xyxiao20@fudan.edu.cn }
\def\thefootnote{\arabic{footnote}}



\def\thefootnote{\arabic{footnote}}

\section{Introduction}
In recent years, deep neural networks, notably CNNs \citep{lecun1998gradient} and ViTs \citep{dosovitskiy2020image}, have achieved remarkable progress in remote sensing data analysis, enhancing both processing efficiency and analytical accuracy. However, existing research mainly focuses on visual feature extraction, while lacking deep semantic parsing and reasoning capabilities \citep{li2024vision}, limiting model applicability in complex scenarios.

With the advancement of Large-Language Models (LLMs), Vision-Language Models (VLMs), through integrating pre-training and instruction tuning, have demonstrated robust zero-shot learning and generalization in multimodal tasks \cite{dai2023instructblip}. This has inspired researchers to explore the deep integration of visual models with LLMs.

Although models designed for optical remote sensing images, like RSGPT \citep{hu2023rsgpt} and GeoChat \citep{kuckreja2024geochat}, have shown preliminary achievements, they struggle to perform well in SAR applications. SAR images inherently pose significant interpretation challenges due to their scattering imaging mechanisms, characterized by blurred target edges, dispersed speckles, and orientation sensitivity. Meanwhile, existing SAR datasets primarily focus on visual recognition tasks \citep{kuckreja2024geochat, cheng2022nwpu, zhang2023rs5m}, leaving a critical shortage of large-scale, high-quality image-text alignment datasets. Both these intrinsic characteristics and data limitations impede the advancement of VLMs in the SAR domain.

\begin{figure*}[th]
    \centering
    \includegraphics[width=0.73\linewidth]{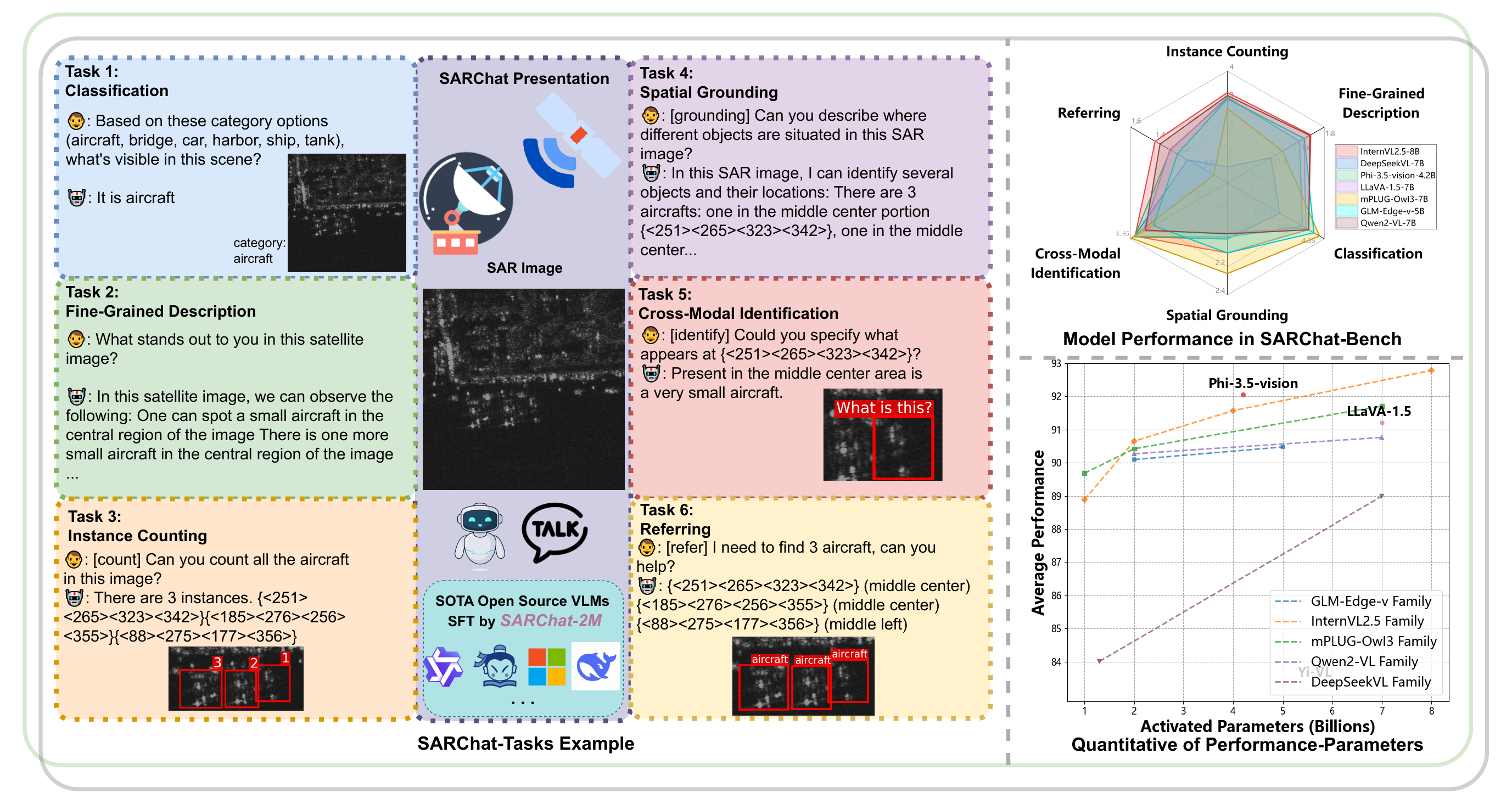}
    \caption{\textbf{An overview of \ourmethod-Bench-2M.} The left figure demonstrates the representative tasks realized with the SAR image-text dataset, \ourmethod-2M, constructed in this paper. Validating the dataset's efficacy and superiority in supporting multi-task applications. The right figure presents the correlation radar charts and quantitative line graphs derived from the performance evaluation of 16 VLMs basing on this dataset, establishing the benchmark (\ourmethod-Bench) within this domain.}
    \label{fig1-overview}
\end{figure*}

Current VLMs are primarily trained on conventional natural images without extensive fine-tuning for the SAR vertical domain. Despite their strong visual capabilities for natural images, these VLMs still have significant room for improvement in SAR image interpretation. Building upon the SARDet-100K dataset \citep{li2024sardet100k} with its rich SAR imagery and detection annotations, we construct \ourmethod-Bench-2M, a task-oriented SAR-specific image-text pair dataset, to address the insufficient SAR image interpretation capabilities of existing VLMs.

As shown in Figure \ref{fig1-overview}, we present \ourmethod-2M, a large-scale multimodal conversational dataset for SAR images, and establish \ourmethod-Bench, a comprehensive multimodal task-oriented benchmark for the SAR domain. The \ourmethod-2M dataset contains approximately 2 million high-quality SAR image-text pairs across maritime, terrestrial, and urban scenarios, featuring fine-grained semantic descriptions and multi-scale resolutions (0.3-10 meters). The dataset supports major vision-language tasks such as image captioning, VQA(Visual Question Answering), visual localization, and object detection. To systematically evaluate model performance in these domains, we design six specific benchmark tasks in \ourmethod-2M: \textbf{classification}, \textbf{fine-grained description}, \textbf{instance counting}, \textbf{spatial grounding}, \textbf{cross-modal identification}, and \textbf{referring}. To validate the effectiveness of our dataset and benchmark, we conduct extensive experiments by fine-tuning 16 state-of-the-art VLMs of varying parameter scales, including InternVL2.5, DeepSeekVL, GLM-Edge-V, and the mPLUG-Owl family. Through training on \ourmethod-2M, these visual language models (VLMs) acquire comprehensive multi-task capabilities in SAR interpretation, as demonstrated by our systematic evaluation on \ourmethod-Bench.

The primary contributions of this paper are as follows:
\begin{enumerate}
\item The construction of \ourmethod-2M, the largest SAR remote sensing instruction-following dataset to date, comprising over 2 million high-quality image-text pairs across multi-scenario task-oriented dialogues, alleviating the knowledge scarcity of VLMs in the SAR domain.

\item The development of \ourmethod-Bench, a comprehensive SAR domain multimodal benchmark encompassing six core tasks (classification, description, counting, localization, recognition, and refering), enabling systematic evaluation of vision-language models through multi-dimensional assessment metrics.

\item It pioneers a research paradigm applicable to the SAR field, providing reference ideas for the construction of models in other remote-sensing vertical domains. The methods and processes adopted in data collection, annotation, as well as model training and evaluation in this study have good generality and extensibility.
\end{enumerate}
\begin{figure*}[h]
    \centering
    \includegraphics[width=0.7\linewidth]{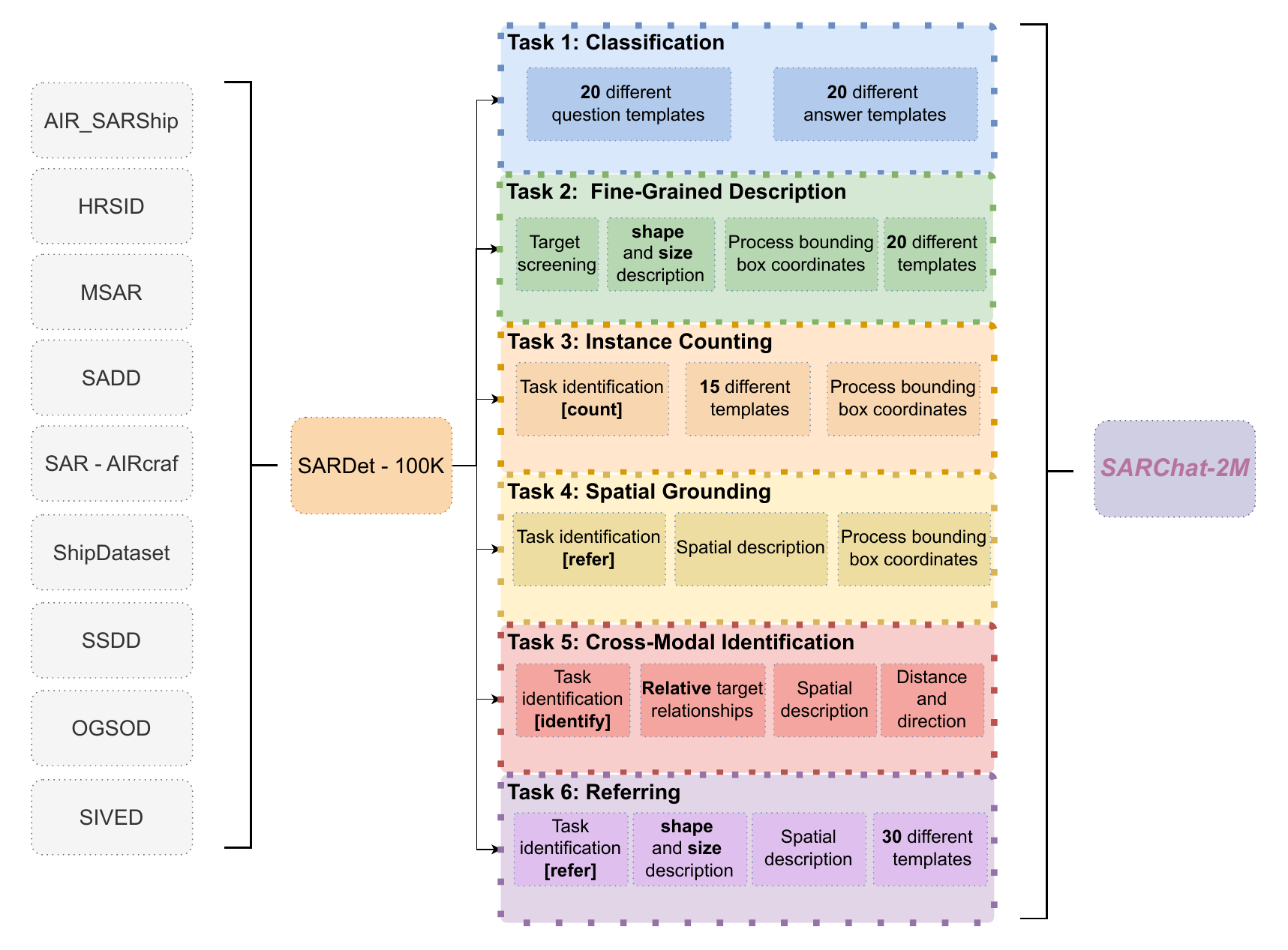}
    \caption{\textbf{Construction of \ourmethod-2M dataset.} On the left, ten existing SAR detection benchmark datasets. The middle part is the SARDet-100K dataset, formed by integrating the ten datasets on the left. On the right, six core tasks constructed based on the dataset are presented, with each task corresponding to different task identifiers, operation steps, and relevant templates.}
    \label{fig2-data}
\end{figure*}

\section{Related Work}

\subsection{VLMs for Remote Sensing}
VLMs are capable of converting images into natural language descriptions and parsing the relationships between objects, demonstrating remarkable performance in tasks such as text-image retrieval, image captioning, and visual question answering. Recently, models like RemoteClip \citep{liu2024remoteclip} have been applied to the field of remote sensing images, primarily focusing on cross-modal retrieval and zero-shot classification. However, these models have not addressed tasks such as image description generation and visual grounding. The RSGPT model has achieved text description and visual question answering for remote sensing images, but it has not expanded to tasks such as classification and detection. The GeoChat model has advanced multi-task conversational processing of high-resolution remote sensing imagery, including scene classification, visual question answering, multi-turn dialogue, visual grounding, and reference object detection. However, these models, including GeoChat, predominantly rely on optical remote sensing training data, leading to suboptimal performance in SAR-specific interpretation tasks. EarthGPT \citep{zhang2024earthgpt} has extended the application of multimodal large language models to the remote sensing field through instruction tuning, but its performance in SAR image multi-task processing still needs improvement. Compared with natural images, the interpretation of SAR images is more challenging, which poses higher demands on the model's processing capabilities and adaptability.

\subsection{Remote Sensing Vision-Language Datasets}
Remote sensing datasets are essential for models that interpret remote sensing imagery. Existing datasets such as UCM Captions \cite{7546397}, Sydney Captions \cite{qu2016deep}, RSICD \citep{lu2017exploring}, RSITMD \citep{yuan2022exploring}, and RSVG \citep{zhan2023rsvg} provide preliminary resources for studying the correlation between remote sensing images and text. However, these datasets are limited not only in scale but also in modality, containing only optical images without SAR data, leaving SAR interpretation capabilities largely unexplored. Although large-scale datasets like MillionAID \citep{long2021creating}, FMoW \citep{christie2018functional}, and BigEarthNet \citep{sumbul2019bigearthnet} exist, they lack text-image pairs. The RS5M  dataset \citep{zhang2023rs5m}, containing 5 million image-text pairs, is still limited to optical images. The MMRS-1M dataset \citep{zhang2024earthgpt}, which covers optical, infrared, and SAR modes, has a very low proportion of SAR image-text data. Therefore, this paper constructs the \ourmethod-2M dataset, which focuses on SAR images and contains over 2 million image-text pairs, covering tasks such as classification, detection, caption generation, VQA, and visual grounding.

\section{Data Construction and Description}

\subsection{The Procedure of Data Construction}

\subsubsection{Dataset Overall}

As shown in Figure \ref{fig2-data}, we propose \ourmethod-2M, a multi-task dataset for SAR images, comprising 2 million multimodal dialogue samples (1,836,912 train and 226,636 test samples) to ensure robust model training and evaluation.

Based on the SARDet-100K dataset \citep{li2024sardet100k}, it incorporates multimodal adaptations and enhanced language annotations from ten established SAR detection benchmarks such as AIR-SARShip(1.0\&2.0) \citep{DBLP:journals/remotesensing/WangWZDW19a}, HRSID \citep{9127939}, MSAR \citep{chen2022large}, SADD \citep{rs13183690}, SAR-AIRcraf \citep{zhirui2023sar}, ShipDataset \citep{wang2019sar}, SSDD \citep{zhang2021sar}, OGSOD \citep{wang2023category}, and SIVED \citep{lin2023sived}. The \ourmethod-2M covers six semantic categories (ships, tanks, bridges, ports, aircraft, and automobiles) and supports six core SAR image analysis tasks: \textbf{classification}, \textbf{fine-grained description}, \textbf{instance counting}, \textbf{spatial grounding}, \textbf{cross-modal identification}, and \textbf{referring}. These diverse tasks are designed to enhance VLMs' capabilities in SAR image interpretation, with 2 million carefully curated annotations for cross-modal learning.




\subsubsection{Task Definition}
Based on the characteristics of SAR images and the core capabilities of the VLM, this study constructs an evaluation system consisting of six tasks. The definitions of each task are as follows:


\noindent\textbf{(1) Classification:} Classification is a fundamental task in SAR image interpretation that evaluates the VLM's basic visual understanding through target category discrimination.

\noindent\textbf{(2) Fine-Grained Description:} The fine-grained description task focuses on both target category identification and geometric attribute analysis in SAR imagery. Beyond basic classification, it evaluates the VLM's capability to extract detailed morphological features and spatial orientations, demonstrating the model's proficiency in reasoning about SAR-specific spatial-geometric relationships.

\noindent\textbf{(3) Instance Counting:} This task requires accurate counting of multiple SAR targets while extracting their spatial coordinates and orientation information. The key challenge lies in preventing double-counting errors, particularly in complex scenes where multiple targets overlap. The model must maintain robust counting performance while handling various target densities and background complexities.

\noindent\textbf{(4) Spatial Grounding:} This task challenges the model to interpret and reason about complex spatial relationships between multiple targets in SAR imagery, including their relative positions, distances, and directional relationships. The key challenge lies in accurately understanding and describing diverse spatial configurations, especially in scenes with multiple interacting objects and varying spatial layouts. The model must demonstrate precise spatial reasoning abilities while handling complex multi-target scenarios and maintaining consistent performance across different scene compositions.

\noindent\textbf{(5) Cross-Modal Identification:} Given specified spatial coordinates, the VLM infers target attributes and generates comprehensive descriptions (size, morphology, direction, distance). This task examines the model's ability to fuse and reason about multimodal information in SAR interpretation.

\noindent\textbf{(6) Referring:} This reverse-reasoning task challenges the model to locate specific instances in SAR images from textual descriptions. The key challenge lies in bridging semantic-visual gaps while accurately determining target spatial orientations, requiring robust cross-modal reasoning capabilities across varied scene configurations.

\subsubsection{Task-Oriented Data Generation}


Based on the characteristics of the six tasks, this study designs a multimodal dialogue data generation scheme. The specific rules and implementation logic are as follows, with detailed templates provided in the Appendix \ref{example}:

\noindent\textbf{(0) Dataset Definitions}

Our dataset adopts a unified representation scheme across all visual-language tasks to ensure consistency and interpretability. The spatial information is uniformly encoded using the bounding box format \{<$x_1$><$y_1$><$x_2$><$y_2$>\}, where ($x_1$,$y_1$) and ($x_2$,$y_2$) denote the top-left and bottom-right coordinates respectively. Spatial relationships are structured through a standard 3×3 grid system (consisting of top-left, top, top-right, left, middle, right, bottom-left, bottom, bottom-right regions).

To explicitly specify different task requirements, we incorporate task-specific prompts: $[count]$ for Instance Counting Task, $[grounding]$ for Spatial Grounding Task, $[identify]$ for Cross-Modal Identification, and $[refer]$ for Referring Task. These prompts help guide the model's attention to the relevant aspects of each task.

These definitions form the foundational framework for our task formulations and evaluation metrics, enabling systematic assessment of visual-language models' capabilities.

\noindent\textbf{(1) Classification Task}

The Classification Task assesses the model's SAR image recognition capabilities through 20 distinct question-answer template pairs. Random template combinations enhance data diversity, with standardized notation for multi-target scenarios.

\noindent\textbf{(2) Fine-Grained Description Task}

Fine-Grained description evaluates the model's structured parsing of satellite imagery through comprehensive quality control. Following our dataset definitions, we filter images below 224×224 pixels and exclude targets with area ratio $R$ < 1\% (Equation \ref{eq:ratio}). Targets with aspect ratios exceeding 10:1 or out-of-bounds coordinates are removed. Size descriptions are categorized using area-ratio thresholds (small: <5\%, large: >30\%). We construct 40 interaction templates to accommodate multi-target scenarios. The calculation of $R$ is formulated as follows:

\begin{equation}
R = \frac{w_{box} \times h_{box}}{W_{img} \times H_{img}} \times 100%
\label{eq:ratio}
\end{equation}

where $w_{box}$ and $h_{box}$ denote the width and height of the target bounding box, respectively; $W_{img}$ and $H_{img}$ represent the width and height of the image.

\noindent\textbf{(3) Instance Counting Task}

As a fundamental component of our visual reasoning system, this task focuses on evaluating the model's quantitative counting capabilities. We designate 15 question templates with $[count]$ identifiers to specify the task requirements, while utilizing our unified bounding box format for structured output representation. The framework supports extended expressions for multi-instance scenarios through coordinate serialization.

\noindent\textbf{(4) Spatial Grounding Task}

Spatial Grounding assessment evaluates the model's proficiency in characterizing structural relationships among multiple target objects. Leveraging our established grid system, we quantify spatial relationships through two primary mechanisms: relative distance metrics (with proximal threshold defined as $(W_{img}+H_{img})/8$) and directional relationships (encompassing horizontal, vertical, and diagonal orientations). The framework incorporates 15 spatial-relationship templates, each prefixed with $[grounding]$ identifiers, conforming to our unified spatial representation scheme.

\noindent\textbf{(5) Cross-Modal Identification} 

Cross-modal parsing evaluation employs a three-tier feature description system. Spatial positioning utilizes a 3×3 grid partitioning scheme for orientation description. Quantitative classification encompasses five-level size descriptions based on area-ratio  $R$ thresholds ($\geq$0.4:very large; $\geq$0.25:large; $\geq$0.1:medium; $\geq$0.03:small; $<$0.03:very small) and morphological analysis through bounding-box aspect ratios ($>$1.5:wide-body; 0.67$\leq$ratio$\leq$1.5:approximately square; $<$0.67:tall-body).

Feature integration combines spatial-size-morphological elements into comprehensive target profiles. The system implements 20 differential response templates with a dedicated $[identify]$ instruction identifier and structured output templates.

\noindent\textbf{(6) Referring Task}

Referring evaluates cross-modal correlation capabilities between natural language and image regions. Queries follow the pattern "Where is the \{category\}?", prefixed with $[refer]$ identifiers. The task outputs both precise bounding box coordinates and grid-based orientation descriptions, adhering to our unified spatial representation framework through nested parenthetical notation.

\subsection{Quantitative Analysis of Datasets} \label{sec:data_construction}
The quantitative analysis in this study focuses on two key dimensions: category distribution and object morphological patterns.

\begin{table}[h]
    \centering
    \small

    \setlength{\heavyrulewidth}{1.5pt}
    \setlength{\lightrulewidth}{1.5pt}
    \begin{tabular}{ccc}
        \toprule
        \textbf{Category}&\textbf{Training}&\textbf{Test}\\ 
        \midrule
        Ship&93,373 (46.98\%)&10,741 (44.38\%)\\
        Aircraft&40,705 (20.48\%)&6,779 (28.01\%)\\
        Car&9,561 (4.81\%)&1,230 (5.08\%)\\
        Tank&24,187 (12.17\%)&1,773 (7.33\%)\\
        Bridge&27,615 (13.89\%)&3,281 (13.56\%)\\
        Harbor&3,306 (1.66\%)&399 (1.65\%)\\
        \bottomrule
    \end{tabular}
        \caption{Category Distribution Statistics}
    \label{tab1-clases}
\end{table}
\textbf{(1) Category Distribution Characteristics}



As shown in Table \ref{tab1-clases}, the ship category dominates both training and test sets (46.98\% and 44.38\% respectively), while the harbor category represents less than 2\%. A significant distribution shift is observed in the aircraft category, with a 7.53\% increase in the test set compared to the training set. Categories such as cars, tanks, and bridges maintain moderate and stable proportions across both sets. This class distribution aligns with real-world SAR imagery characteristics, where certain target types naturally appear more frequently than others due to the inherent nature of SAR remote sensing applications and operational scenarios.


\textbf{(2) Object Morphology Analysis}




This study quantify geometric characteristics using aspect ratio (AR):
\begin{equation}
    AR = \frac{h_{box}}{w_{box}} \times 100\%
\end{equation}

\newcolumntype{C}{>{\centering\arraybackslash}X}
\begin{table}[ht]
    \centering
    \small 
    \setlength{\heavyrulewidth}{1.5pt}
    \setlength{\lightrulewidth}{1.5pt}
    \begin{tabularx}{\linewidth}{lCCC}
        \toprule
        Metric & Training Set & Test Set & Diff-Rate \\
        \midrule
        Mean & 1.28 & 1.26 & -0.02 \\
        Median & 1.062 & 1.05 & -0.017 \\
        SD & 1.18 & 0.91 & -0.22 \\
        \bottomrule
    \end{tabularx}
        \caption{Aspect Ratio Distribution Comparison}
    \label{tab2-aspect_ratio}
\end{table}

As shown in Table \ref{tab2-aspect_ratio}, the differences in central tendency between training and test sets are minimal (mean: -0.02, median: -0.017). The test set exhibits a 0.22 lower standard deviation, indicating a more concentrated distribution. The key morphological distribution intervals of targets are illustrated in Appendix \ref{Morphological}.

The dataset exhibits three distinct morphological categories based on aspect ratio (AR): broad-bodied (AR $\leq$ 0.67), nearly square-shaped (0.67 $<$ AR $\leq$ 1.5), and tall-bodied (AR $>$ 1.5). Detailed distribution analysis can be found in Appendix \ref{appendix:analys-mor}.

\begin{table*}[htbp]
\scriptsize
\setlength{\tabcolsep}{2.5pt}

\centering
\begin{tabular}{ccccccccccccccccc}
\Thickline
\multirow{3}{*}{\textbf{Model}} & \multirow{3}{*}{\textbf{Param}} & \multirow{3}{*}{\begin{tabular}[c]{@{}c@{}}\textbf{Avg}\\ \textbf{score}\end{tabular}} & \multicolumn{14}{c}{\textbf{Tasks}} \\
\cline{4-17}
 &  &  & \multirow{2}{*}{\begin{tabular}[c]{@{}c@{}}\textbf{Only}\\ \textbf{count}\end{tabular}} & \multicolumn{2}{c}{\textbf{Instance Count}} & \multirow{2}{*}{\begin{tabular}[c]{@{}c@{}}\textbf{Abstract}\\ \textbf{position}\end{tabular}} & \multicolumn{2}{c}{\textbf{Spatial Ground}} & \multicolumn{2}{c}{\textbf{Cross-Modal ID}} & \multicolumn{2}{c}{\textbf{Multi-target Ref}} & \multicolumn{2}{c}{\textbf{Single-target Ref}} & \multirow{2}{*}{\textbf{Descript}} & \multirow{2}{*}{\textbf{Class}} \\
\cline{5-6} \cline{8-9} \cline{10-11} \cline{12-13} \cline{14-15}
 &  &  &  & \textbf{IoU=.25} & \textbf{IoU=.5} &  & \textbf{Multi} & \textbf{Single} & \textbf{Multi} & \textbf{Single} & \textbf{IoU=.25} & \textbf{IoU=.5} & \textbf{IoU=.25} & \textbf{IoU=.5} &  &  \\
\Thickline
\multirow{4}{*}{\textbf{InternVL2.5}} & \textbf{8B} & 92.79 & \textbf{74.14} & 61.37 & 52.17 & 81.25 & 62.25 & 87.91 & 98.84 & 98.98 & 37.49 & 23.46 & 74.86 & 60.13 & 63.43 & 97.25 \\
 & \textbf{4B} & 91.57 & 72.68 & 57.54 & 47.35 & 83.33 & 60.89 & 85.90 & 98.01 & 98.76 & 34.05 & 18.86 & 69.92 & 55.29 & 58.84 & 97.27 \\
 & \textbf{2B} & 90.55 & 71.52 & 54.11 & 44.22 & 50.00 & 60.81 & 81.92 & 97.79 & 98.63 & 27.05 & 13.91 & 68.50 & 52.16 & 56.36 & 96.69 \\
 & \textbf{1B} & 88.89 & 69.87 & 50.18 & 39.35 & 0.00 & 56.30 & 82.24 & 96.98 & 98.60 & 22.13 & 9.94 & 62.33 & 44.99 & 53.30 & 96.65 \\
\hline
\multirow{2}{*}{\textbf{DeepSeekVL}} & \textbf{7B} & 88.99 & 20.66 & 8.49 & 4.19 & 64.29 & 65.32 & 85.78 & 98.97 & 99.05 & 28.75 & 13.66 & 64.34 & 48.84 & 51.08 & 93.23 \\
 & \textbf{1.3B} & 84.01 & 19.61 & 4.00 & 1.32 & 75.00 & 60.38 & 82.00 & 96.40 & 97.45 & 16.11 & 6.23 & 53.58 & 34.28 & 44.44 & 47.37 \\
\hline
\textbf{Phi-3.5-vision} & \textbf{4.2B} & 92.06 & 72.69 & 57.48 & 47.60 & 62.50 & 58.85 & 87.29 & 98.93 & 98.59 & 31.65 & 17.16 & 70.95 & 55.70 & 59.95 & 96.42 \\
\hline
\multirow{2}{*}{\textbf{GLM-Edge-V}} & \textbf{2B} & 90.20 & 71.59 & 51.97 & 40.37 & 42.86 & 59.15 & 86.33 & 97.54 & 98.60 & 24.15 & 10.66 & 65.57 & 46.46 & 57.86 & 97.39 \\
 & \textbf{5B} & 90.48 & 73.44 & 56.30 & 44.56 & 75.00 & 61.38 & 89.96 & 96.69 & 95.96 & 30.68 & 15.41 & 69.36 & 51.81 & 61.45 & 98.02 \\
\hline
\multirow{3}{*}{\textbf{mPLUG-Owl3}} & \textbf{7B} & 91.71 & 71.00 & 48.07 & 35.27 & \textbf{100.00} & 56.37 & 93.32 & \textbf{99.27} & \textbf{99.51} & 19.72 & 7.66 & 57.27 & 38.00 & 54.65 & 98.80 \\
& \textbf{2B} & 90.32 & 67.56 & 41.56 & 28.83 & 75.00 & 45.65 & 97.58 & 98.95 & 99.42 & 14.91 & 5.42 & 50.46 & 30.16 & 41.76 & 98.31 \\
 & \textbf{1B} & 89.68 & 67.03 & 38.64 & 24.98 & 75.00 & 44.07 & 97.19 & 98.72 & 98.87 & 11.86 & 4.12 & 44.34 & 24.02 & 40.16 & 98.06 \\
\hline
\multirow{2}{*}{\textbf{Qwen2-VL}} & \textbf{7B} & 90.76 & \textbf{72.79} & 58.51 & 50.24 & 0.00 & 64.17 & 83.87 & 97.54 & 99.18 & 39.11 & 26.29 & 70.55 & 57.04 & 63.11 & 97.30 \\
 & \textbf{2B} & 90.27 & 69.63 & 53.62 & 45.47 & 50.00 & 59.04 & 78.49 & 97.55 & 99.26 & 32.60 & 20.12 & 65.31 & 51.53 & 55.20 & 96.88 \\
\hline
\textbf{LLaVA-1.5} & \textbf{7B} & 91.21 & 71.89 & 56.89 & 46.80 & 57.14 & 62.70 & 85.79 & 97.84 & 98.42 & 30.81 & 15.48 & 71.89 & 56.70 & 61.35 & 96.90 \\
\hline
\textbf{Yi-VL} & \textbf{6B} & 84.35 & 32.62 & 14.35 & 9.44 & 75.00 & 53.68 & 72.38 & 93.63 & 97.95 & 7.76 & 2.69 & 32.95 & 16.63 & 38.15 & 95.32 \\
\Thickline
\end{tabular}
\caption{Performance comparison of different vision-language models}
\label{tab3-vl-models}
\end{table*}

\section{SARChat-Bench Evaluation Method and Settings} 

This section details the evaluation methodology of \ourmethod-Bench, a standardized benchmark suite we designed for comprehensive assessment of VLMs in SAR interpretation. The benchmark covers six fundamental tasks that evaluate the model's core capabilities across information processing, target localization, and semantic understanding, providing multi-dimensional insights into visual-language model performance in the SAR domain. The evaluation framework ensures fair and thorough assessment of VLMs' capabilities across different SAR interpretation scenarios.

\subsection{Evaluation Metrics}




\noindent\textbf{(1) Accuracy:} A core metric reflecting model prediction fit, calculated as:

\begin{equation}
Acc = \frac{TP}{TP + FP + FN} \times 100\%
\end{equation}

where $TP$ denotes correct positive predictions, $FP$ represents false positive predictions, and $FN$ indicates false negative predictions.

\noindent\textbf{(2) Intersection over Union (IoU):} In tasks involving localization, identification, and reference, IoU is a key metric measuring the overlap between predicted and ground-truth bounding boxes (bbox). Higher IoU values indicate greater overlap and better localization performance. All IoU-related calculations in this paper are performed with thresholds of 0.25 and 0.5.

\noindent\textbf{(3) Overall Score Calculation:} 

\begin{equation}
S_m = \sum_{t \in T} a_{m,t} \times \frac{n_t}{\sum_{i \in T} n_i}
\end{equation}

Among them, $n_t$ represents the sample size of task $t$, $a_{m,t}$ denotes the accuracy of model $m$ on task $t$, and $T$ is the set of all tasks. The detailed calculation of $a_{m,t}$ for each task can be found in Appendix \ref{task-cal}.


\subsection{Assessment Methods}

This section elaborates on the specific evaluation method processes for six types of tasks.

\noindent(1) \textbf{Instance Counting:} Compare predicted and label object counts for single-class evaluation, where counting accuracy is measured by $Acc$ and object localization precision is evaluated using Intersection over Union ($IoU$).

\noindent\textbf{(2) Spatial Grounding:} Evaluate spatial accuracy through $IoU$-based bbox matching and abstract position analysis (e.g., "top", "bottom") from natural language descriptions.

\noindent\textbf{(3) Cross-Modal Identification:} Calculate $IoU$ between predicted and ground-truth bboxes for both single and multiple target scenarios to assess cross-modal matching capability.

\noindent\textbf{(4) Referring:} Assess referring accuracy through $IoU$ metrics in both single-target and multi-target contexts.

\noindent\textbf{(5) Fine-Grained Description:} Segment predictions and ground-truth into short phrases, extract category and position information, and compare content sets for detailed description evaluation.

\noindent\textbf{(6) Classification:} Compare predicted and ground-truth categories to assess classification accuracy $Acc$.

\section{Experiments and Analysis}

\subsection{Implementation Details}



During the fine-tuning stage, the model is trained for 1 epoch with a batch size of 4 (effective batch size 32 with gradient accumulation steps of 4). This study employs LoRA training method with rank=8 and alpha=32 on all linear layers. The learning rate is initialized at 1e-4 with a 0.1 warm-up ratio. All experiments are conducted on 2 NVIDIA A100 GPUs using bfloat16 precision. Training is implemented using the MS-SWIFT framework \citep{zhao2024swiftascalablelightweightinfrastructure} for efficient distributed training. In the most time-consuming case, the training process for a single model took up to 192 hours to complete.

\subsection{Benchmark Evaluation}

To verify the effectiveness and practicality of the \ourmethod-2M dataset, we conducted extensive experiments on SAR image interpretation tasks using sixteen mainstream visual-language models. Our preliminary analysis reveals that these models, despite their strong performance on natural images, struggle significantly with SAR image interpretation without domain-specific fine-tuning, demonstrating the critical importance of SAR-domain adaptation. Specifically, we conducted a detailed before-and-after analysis on InternVL2-8B, which achieved the best performance among all tested models, to quantitatively demonstrate the impact of SAR fine-tuning. The results are presented in Appendix~\ref{ftornot}.

As shown in Table \ref{tab3-vl-models}, the evaluated models include recent advances such as Qwen2-VL\citep{wang2024qwen2}, InternVL2.5\citep{chen2024expanding}, DeepSeekVL\citep{lu2024deepseekvl}, Phi-3.5-vision\citep{abdin2024phi}, GLM-Edge-V\citep{glm2024chatglm}, mPLUG-Owl3\citep{ye2024mplug}, Yi-VL\citep{young2024yi} and LLaVA-1.5\citep{liu2023llava}. In experiments of different tasks, each model exhibits diverse characteristics:

\noindent\textbf{(1) Instance Counting} requires VLMs to identify the number of specific objects in the image. Two leading model families achieve state-of-the-art performance: InternVL2.5 and Qwen-VL2, reaching accuracies of 74.14\% and 72.79\% respectively. However, the accuracy of most other models fall below 60\%, highlighting both the challenging nature of the counting task and the dataset's effectiveness in differentiating model capabilities.


\noindent\textbf{(2) Spatial Grounding} evaluates models' capability in spatial localization. For abstract position descriptions, mPLUG-Owl3-7B achieves 100\% accuracy, significantly outperforming other models. The mPLUG-Owl3 family maintains superior performance (>90\%) in single-target localization, while other models achieve 80\%-85\%. However, in multi-target scenarios, most models' accuracy drops to approximately 60\%. These results suggest that accurate multi-target spatial information processing remains a crucial area for future model improvements.

\begin{figure*}[th]
    \centering    \includegraphics[width=0.73\linewidth]{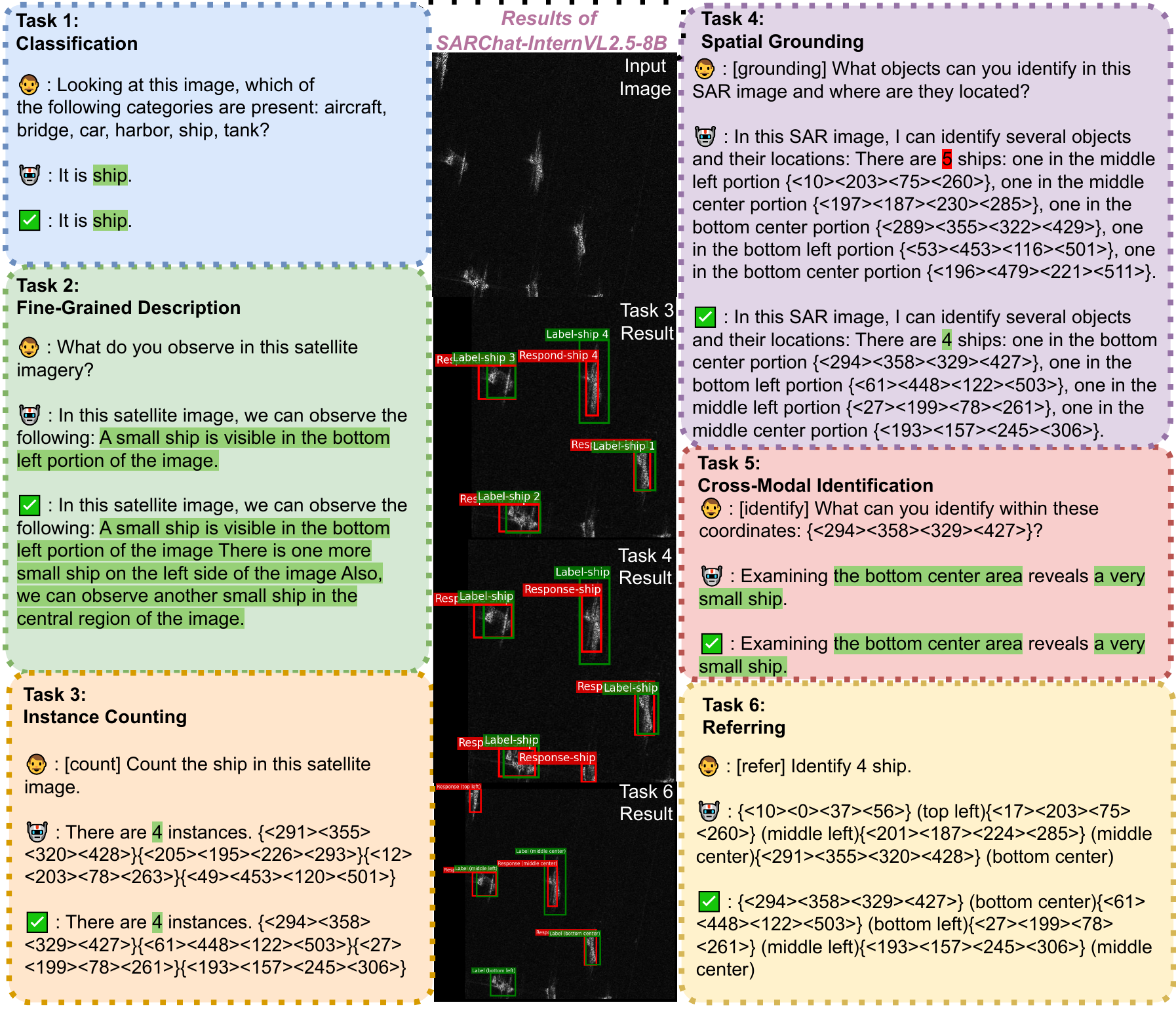}
    \caption{\textbf{Evaluation examples on \ourmethod-Bench.} VLM predictions are shown in green/red for correct/incorrect descriptions, with the ground truth in green and the predictions in red boxes. And [Human], [Bot], and [Check] icons denote user input, VLMs response, and standard output, respectively.}
    \label{fig9-sample}
\end{figure*}

\noindent\textbf{(3) Cross-Modal Identification} focuses on the model's ability to build precise connections between visual information and other modal information. In this experiment, the process from image recognition to text description is mainly concerned. The experimental data shows that for both single-target and multi-target tasks, the accuracy rates of most models exceed 90\%. Among them, the mPLUG-Owl3-7B model performs the best, with the accuracy rates of single-target and multi-target tasks reaching 99.27\% and 99.51\% respectively, fully demonstrating the powerful capabilities of large language models in cross-modal identification tasks.


\noindent\textbf{(4) Referring} challenges models to precisely locate objects in SAR images based on textual descriptions. Our experiments reveal significant performance gaps: models achieve less than 75\% accuracy on single-target tasks and below 40\% on multi-target scenarios. These results highlight the current limitations in cross-modal alignment, particularly in establishing precise text-to-object correspondences within SAR imagery.

\noindent\textbf{(5) Fine-Grained Description} requires the model to provide detailed feature and attribute descriptions of the objects in the image. The experiment shows that the model accuracy rates are in the range of 40\%-63\%. Among them, models with larger parameters such as Qwen2-VL-7B and InternVL2.5-8B perform outstandingly and can give more detailed and accurate descriptions. In contrast, other models with smaller parameter sizes perform poorly, indicating that the accuracy rate of the fine-grained description task is highly sensitive to the model's parameter size.


\noindent\textbf{(6) Classification} evaluates models' ability to categorize images based on their content. According to the table data, regardless of parameter size, series such as InternVL2.5, mPLUG-Owl3, Qwen2-VL, and several other models achieve accuracy rates exceeding 96\%. The performance of these VLMs demonstrates competitiveness with traditional vision classification models.

\noindent \textbf{Summary:} We benchmark 16 mainstream VLMs on \ourmethod-Bench. Model size strongly affects fine-grained description performance but shows little impact on classification. While large models excel in cross-modal and class identify tasks and basic spatial grounding, they struggle with referring, counting, detailed descriptions, and multi-target spatial relationships.

\subsection{Edge-side models for SAR Applications}

This study has multiple edge-side models ($\leq$5B parameters) trained on \ourmethod-2M and evaluates their performances. According to Table \ref{tab3-vl-models}, it demonstrates that these models exhibit task-specific performance variations, achieving remarkable accuracy in cross-modal identification, while showing potential for improvement in referring tasks. These models support domain-specific fine-tuning for rapid task adaptation. After optimization, they can operate efficiently on satellite or ground-edge devices, enabling real-time SAR data processing while reducing dependence on cloud infrastructure and minimizing data transmission costs.

\subsection{Dialogue Visualizaion}


Figure \ref{fig9-sample} presents examples across six tasks from \ourmethod-Bench. The model completes these tasks with reasonable performance, and its coordinate predictions align with the ground truth annotations. In the spatial grounding task, the model identifies an additional ship that was not included in the original annotations, suggesting its potential in detecting previously unmarked targets in SAR imagery.

\section{Conclusion}

This research introduces \ourmethod-2M, a large-scale dataset of two million annotated SAR image-text pairs, addressing the scarcity of language-vision data in the SAR domain. The accompanying \ourmethod-Bench provides a systematic evaluation framework for assessing VLMs in SAR interpretation tasks, facilitating domain-specific knowledge integration and accelerating the development of SAR-oriented VLMs.

\newpage

\section*{Limitation}
Despite the comprehensive scale of \ourmethod-2M based on SARDet-100K dataset, the inherent annotation inconsistencies across different SAR data sources may lead to potential limitations. The varying annotation quality could result in missing targets or imprecise target delineation. Notably, there exist cases where VLMs successfully identify valid targets that were not originally annotated in the dataset, highlighting the annotation completeness challenge in the current benchmark construction.

\section*{Ethics Statement}
In this study, all SAR datasets and methodologies are used strictly for academic research purposes, adhering to their respective licenses and data usage agreements. While our research aims to advance the fundamental understanding of SAR image interpretation, we acknowledge that these technologies could potentially be applied to military or defense-related purposes. We emphasize that the responsible application of such technologies is crucial, and their deployment should strictly comply with relevant regulations and ethical guidelines. The research community should maintain ongoing discussions regarding the dual-use nature of SAR technologies to ensure their development serves beneficial purposes while minimizing potential misuse.

\newpage



\bibliography{custom}


\clearpage
\appendix
\addcontentsline{toc}{chapter}{Appendix}
\section*{Appendix}

\section{Data structure analysis}

\subsection{Word Frequency Analysis of SARChat-2M}
As shown in Figure 4 , location words (such as center, middle, top) and target object words (such as ship, aircraft, tank) have the highest occurrence frequencies in SAR image descriptions, and the adjective "small" is the most frequently used descriptive word.
\begin{figure}[h]
    \centering
    \includegraphics[width=0.8\linewidth]{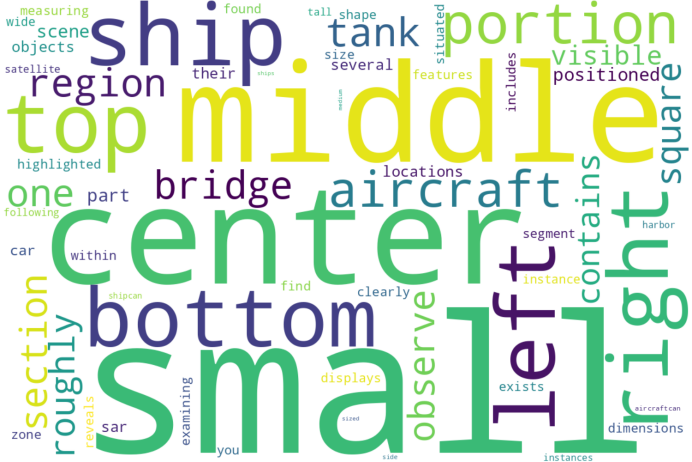}
    \caption{Cloud Map of Word-frequency Distribution }
    \label{fig3-word}
\end{figure}


\begin{figure}[h]
    \centering
    \includegraphics[width=1.0\linewidth]{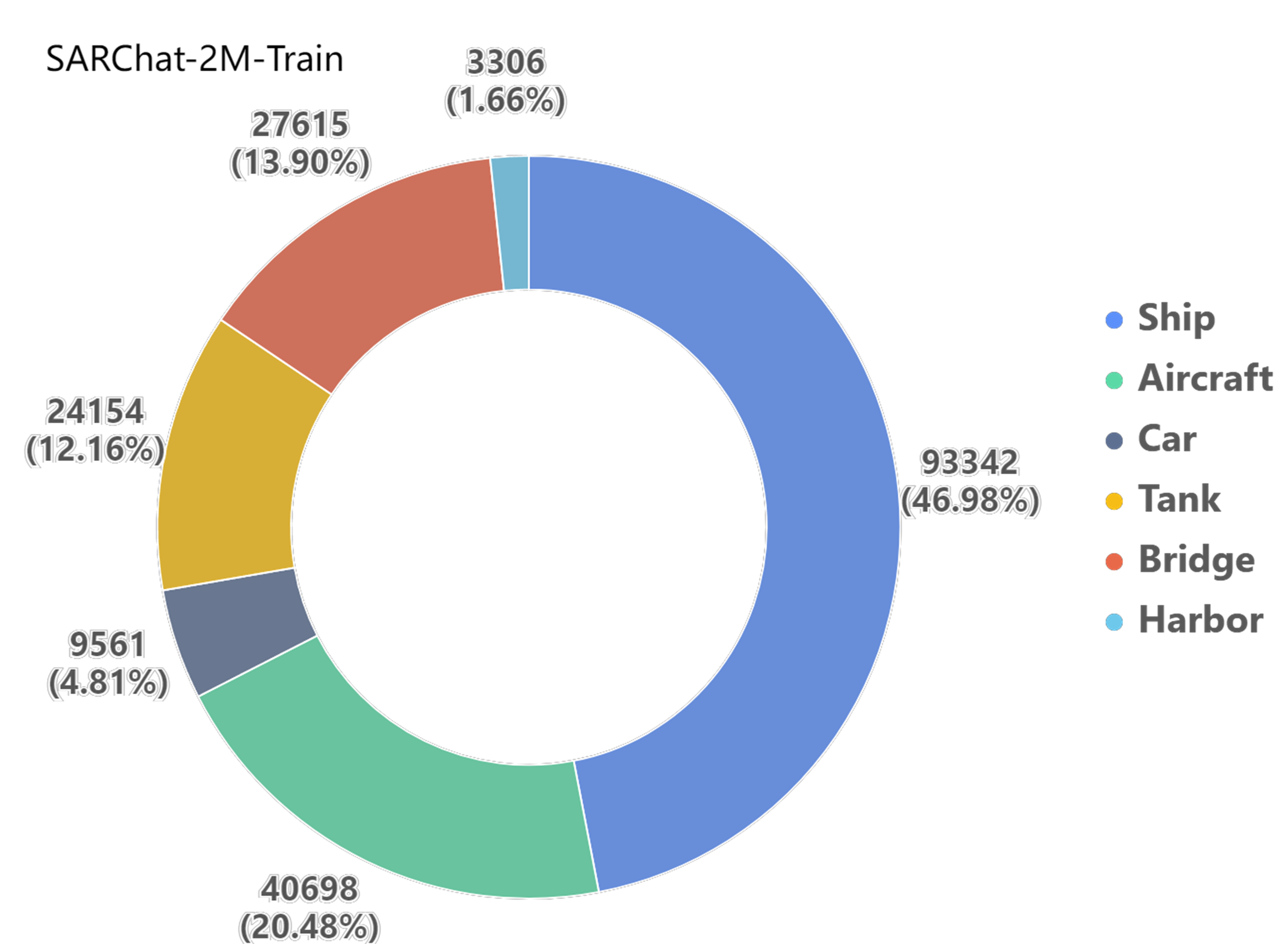}
    \caption{The Proportion Distribution of Samples in the Training Set}
    \label{fig4-train-classes}
\end{figure}
\begin{figure}[h]
    \centering
    \includegraphics[width=1.0\linewidth]{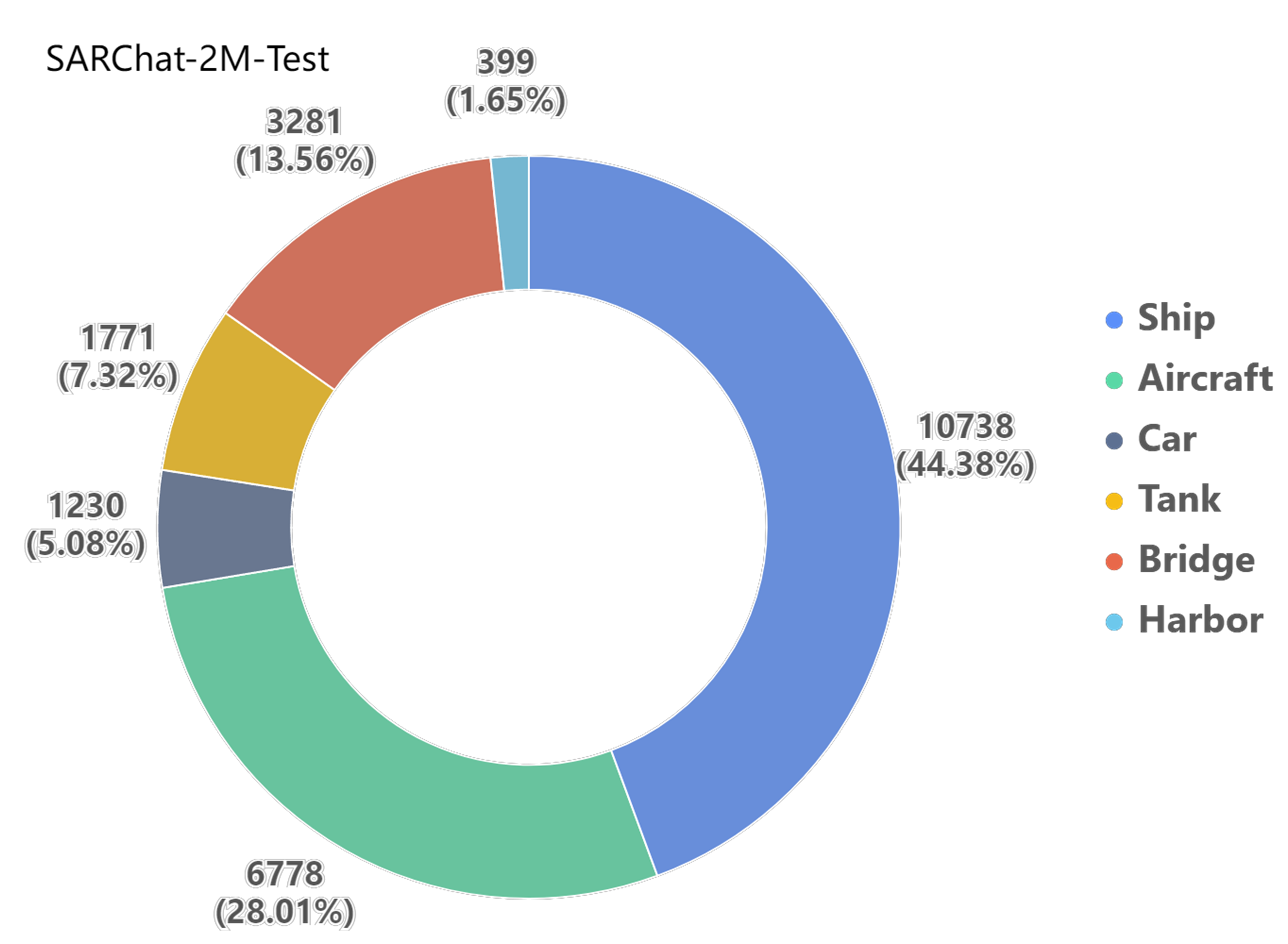}
    \caption{The Proportion Distribution of Samples in the Testing Set}
    \label{fig5-test-classes}
\end{figure}
\subsection{Analysis of Datasets Composition}
\label{Morphological}

\noindent\textbf{Training Set Category Distribution}

As shown in Figure \ref{fig4-train-classes}, in the \ourmethod-2M training set, the category distribution shows significant differences. Among them, the "Ship" category has the largest proportion, reaching 46.98\%, followed by the "Aircraft" category, with a proportion of 20.48\%. These two categories account for the majority of the samples in the training set. It can be seen that the sample distribution in the training set is imbalanced, and the "Ship" and "Aircraft" categories dominate. This may enable the model to learn the features of these two categories more comprehensively during the training process. However, since the samples of other categories are relatively few, the model's ability to learn and generalize their features may be affected to a certain extent.

\noindent\textbf{Test Set Category Distribution}

As shown in Figure \ref{fig5-test-classes}, in the \ourmethod-2M test set, the "Ship" category has the largest proportion among all categories, reaching 44.38\%, and the "Aircraft" category ranks second, with a proportion of 28.01\%. The distribution trends of these two major categories in the test set are similar to those in the training set. This indicates that the test set has a certain similarity to the training set in terms of the overall category distribution and can be used to test the model's generalization ability on data with a similar distribution. However, the slight differences in the proportions also remind us to comprehensively consider various factors when evaluating the model's performance.

\noindent\textbf{Cross-modal Shape Distribution Analysis}

\begin{figure}[h]
\centering
\includegraphics[width=0.9\linewidth]{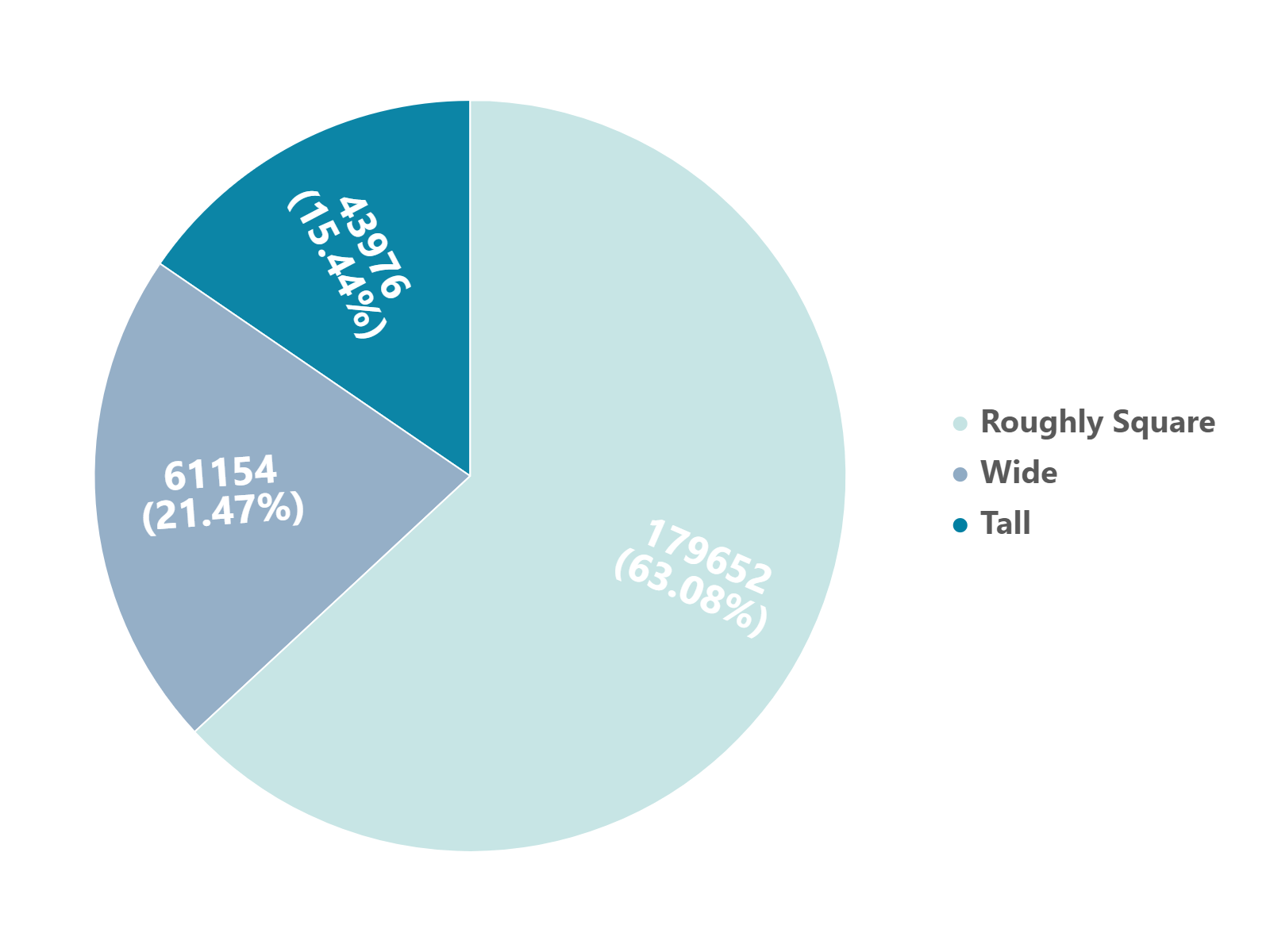}
\small
\caption{Morphological distribution}
\label{fig6-cross_model}
\end{figure}

As shown in Figure \ref{fig6-cross_model}, in the cross-modal identification shape distribution, the "Roughly Square" shape has the largest proportion, with a quantity of 179,652. This shape has an absolute advantage among all shape categories. This means that in the cross-modal identification task, the number of samples of the "Roughly Square" shape is much larger than that of other shapes. The model may be more sensitive to this shape and tend to identify the target as the "Roughly Square" shape during the recognition process. Therefore, when training and optimizing the model, attention should be paid to improving the recognition ability of other shapes to achieve a more balanced recognition effect.

\begin{table*}[h]
    \small
    \setlength{\tabcolsep}{2.5pt}
    \setlength{\arrayrulewidth}{2.5pt} 
    \centering
 
    \begin{tabular}{ccccccccr}
        \toprule
        \multirow{2}{*}{Category} & \multirow{2}{*}{Dataset} & \multirow{2}{*}{Total Samples} & \multirow{2}{*}{Mean AR} & \multirow{2}{*}{Median AR} & \multirow{2}{*}{Std Dev} & \multicolumn{3}{c}{AR Distribution (\%)}\\
        \cmidrule(lr){7 - 9}
        &  &  &  &  &  & AR$\leq$0.67 & 0.67<AR$\leq$1.5 & AR>1.5 \\
        \midrule
        \multirow{2}{*}{Ship} & Train & 93,342 & 1.34 & 1.07 & 1.24 & 28.37 & 39.67 & 31.96 \\
        & Test & 10,738 & 1.308 & 1.026 & 1.10 & 29.34 & 39.82 & 30.84 \\
        \midrule
        \multirow{2}{*}{Aircraft} & Train & 40,698 & 1.074 & 1.047 & 0.32 & 5.85 & 87.67 & 6.48 \\
        & Test & 6,778 & 1.08 & 1.041 & 0.31 & 4.56 & 87.36 & 8.08 \\
        \midrule
        \multirow{2}{*}{Car} & Train & 9,561 & 1.23 & 1.08 & 0.56 & 13.18 & 60.07 & 26.75 \\
        & Test & 1,230 & 1.21 & 1.07 & 0.53 & 12.28 & 62.11 & 25.61 \\
        \midrule
        \multirow{2}{*}{Tank} & Train & 24,15 & 1.10 & 1.00 & 0.84 & 1.58 & 94.29 & 4.13 \\
        & Test & 1,771 & 1.09 & 1.00 & 0.29 & 1.41 & 94.36 & 4.23 \\
        \midrule
        \multirow{2}{*}{Bridge} & Train & 27,615 & 1.56 & 1.18 & 1.92 & 18.38 & 44.83 & 36.79 \\
        & Test & 3,281 & 1.568 & 1.2 & 1.24 & 18.01 & 44.59 & 37.4 \\
        \midrule
        \multirow{2}{*}{Harbor} & Train & 3,306 & 1.20 & 1.01 & 0.72 & 14.19 & 67.93 & 17.88 \\
        & Test & 399 & 1.23 & 1.01 & 0.81 & 15.04 & 68.42 & 16.54 \\
        \bottomrule
    \end{tabular}
       \caption{Analysis of Aspect Ratio of Different Types of Targets}
    \label{tab:analysis-aspect-ratio}
\end{table*}
\noindent\textbf{Distribution Analysis of Morphological Categories}

\label{appendix:analys-mor}
As shown in Table \ref{tab:analysis-aspect-ratio}, nearly square-shaped morphology dominates both datasets, accounting for 39.67\% in training and 59.37\% in test sets, indicating its prevalence in target morphologies. Broad-bodied shapes maintain stable distributions (18.14\% training, 17.72\% test), while tall-bodied shapes show a moderate decrease from training (31.96\%) to test (22.91\%) sets. This distribution diversity enhances the model's generalization capability, though the significant increase in nearly square-shaped samples in the test set demands particular attention during model optimization.

\noindent\textbf{Category-Specific Morphological Patterns}

As shown in Table \ref{tab:analysis-aspect-ratio}, each category displays distinctive morphological characteristics. Bridges exhibit the highest average aspect ratio (1.56) with balanced distribution across all morphologies (18\% broad, 45\% square, 37\% tall). Ships demonstrate diverse shapes (28\% broad, 40\% square, 32\% tall), reflecting their real-world variability. Tanks and aircraft show highly concentrated distributions, with nearly square shapes dominating at 94\% and 87\% respectively, facilitating efficient model learning. Cars and ports maintain moderate distributions with 60-68\% nearly square shapes and balanced remaining proportions.

\subsection{Analysis of Task based on Datasets Composition}

Our dataset is designed to support unified multi-modal tasks through a comprehensive task taxonomy, as shown in Table \ref{tab:task_distribution}. It encompasses six primary tasks: Classification, Fine-Grained Description, Instance Counting, Spatial Grounding, Cross-Modal Identification, and Referring. Among these, the first three tasks are target-quantity independent, while Spatial Grounding, Cross-Modal Identification, and Referring are further categorized into single-object and multi-object variants. This systematic organization enables diverse training scenarios and enhances model generalization capabilities.

\begin{table}[htbp]
    \centering
  
    \resizebox{1.0\linewidth}{!}{
    \begin{tabular}{lcc}
    \noalign{\hrule height 1.5pt}
    Task Type & Train & Test \\
    \noalign{\hrule height 1.5pt}
    Instance Counting & 95493 (5.2\%) & 11794 (5.2\%) \\
    Spatial Grounding & 94456 (5.1\%) & 11608 (5.1\%) \\
    Cross - Modal Identification & 1423548 (77.5\%) & 175565 (77.4\%) \\
    Referring & 95486 (5.2\%) & 11703 (5.2\%) \\
    Fine - Grained Description & 46141 (2.5\%) & 6032 (2.7\%) \\
    Classification & 81788 (4.5\%) & 10024 (4.4\%) \\
    \noalign{\hrule height 1.5pt}
    \end{tabular}
    }
      \caption{Task type distribution in training and test sets}
    \label{tab:task_distribution}
\end{table}

\begin{figure}[h]
    \centering
    \includegraphics[width=0.85\linewidth]{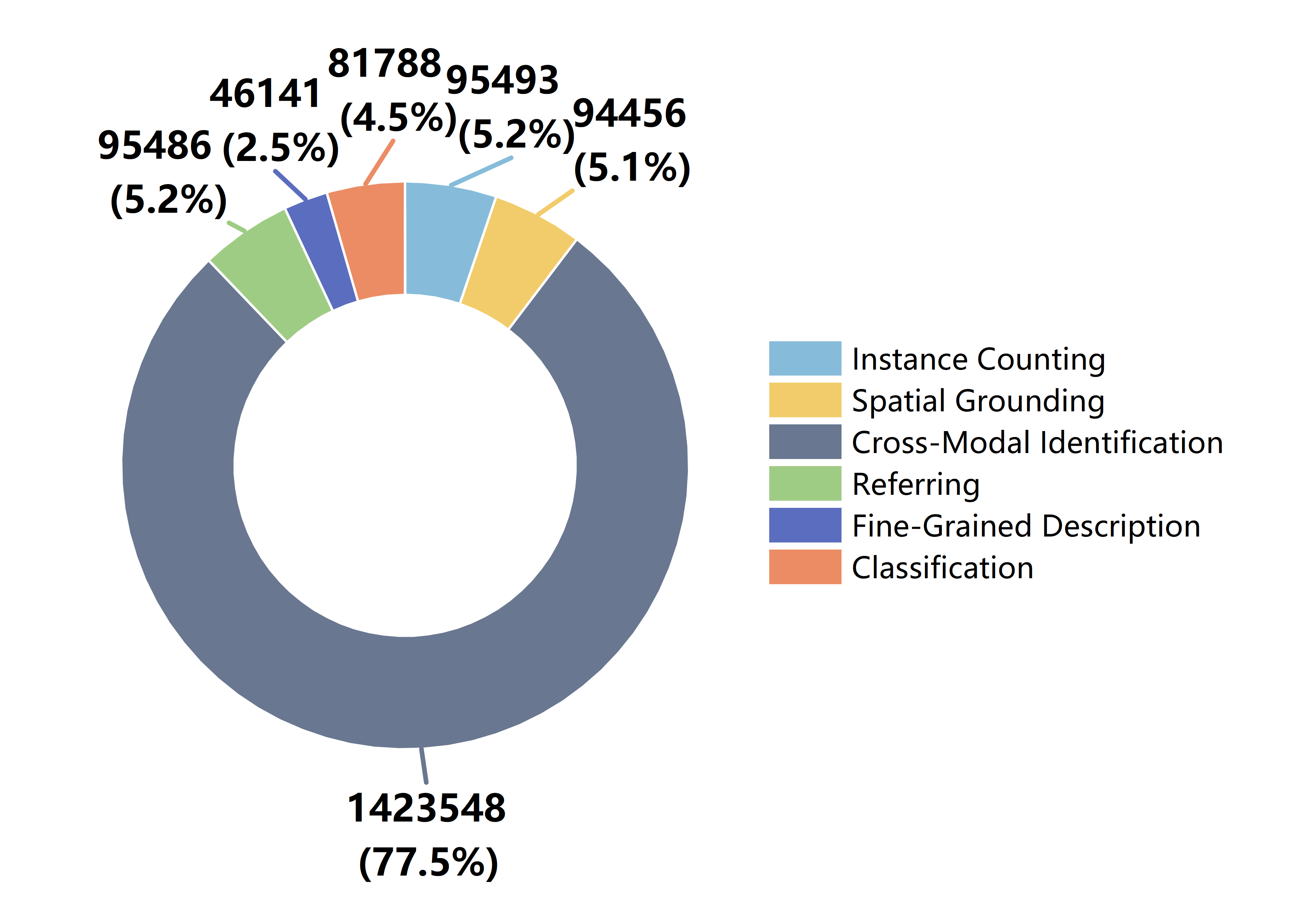}
    {\raggedright
    \caption{Train Task Distribution}
    \label{fig7-task-train}
    }
\end{figure}
 
 \begin{figure}[h]
    \centering
    \includegraphics[width=0.85\linewidth]{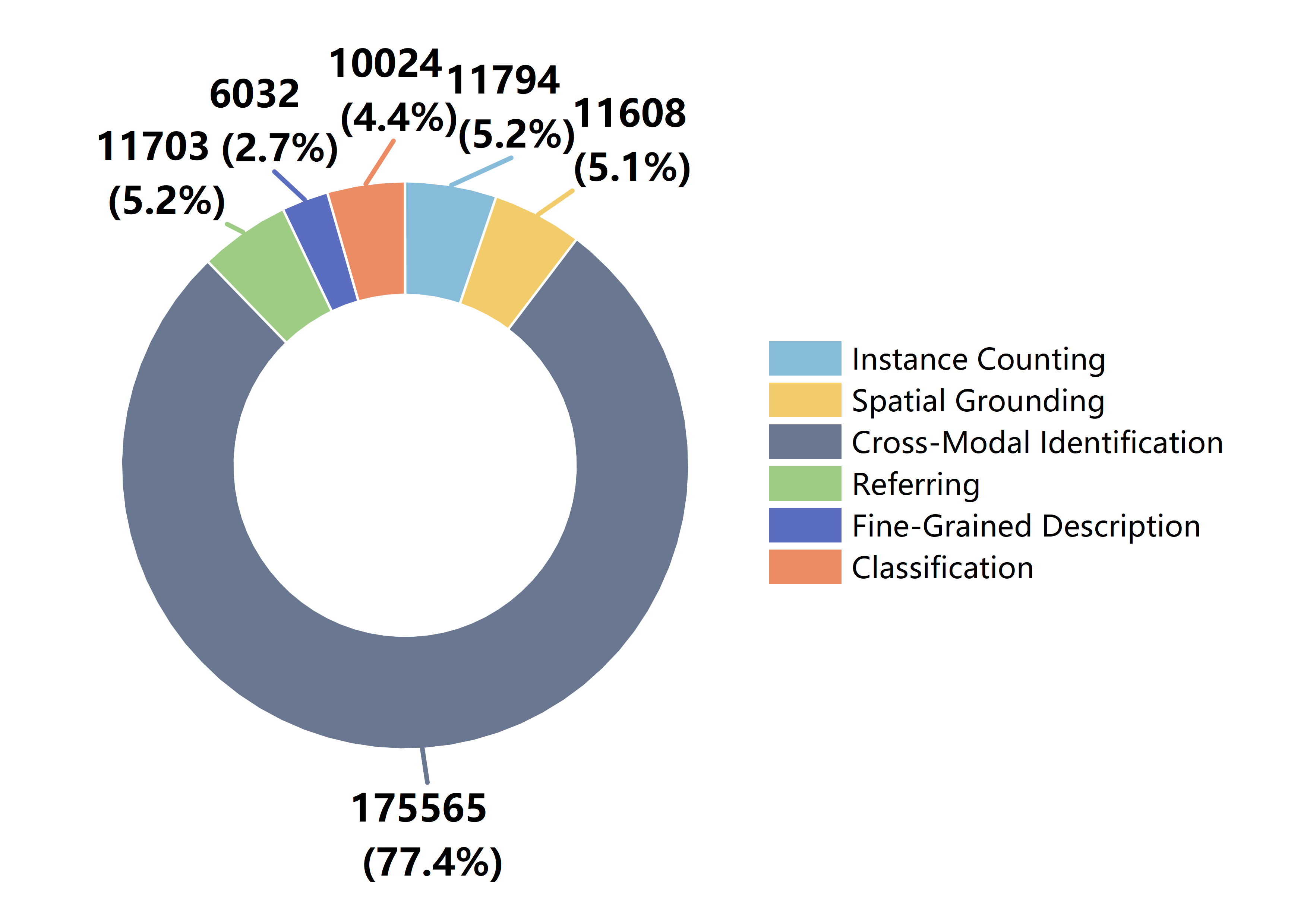}
    \caption{Test Task Distribution}
    \label{fig8-task-test}
\end{figure}

As illustrated in Figure \ref{fig7-task-train}, the training set comprises 1,836,912 entries. Cross-Modal Identification dominates with 1,423,548 entries (77.50\%), enabling robust cross-modal feature learning. Instance Counting and Referring tasks contain 95,493 (5.20\%) and 95,486 (5.20\%) entries respectively, while Spatial Grounding accounts for 94,456 entries (5.14\%). Fine-Grained Description includes 46,141 entries (2.51\%), with its relatively limited data volume potentially affecting model performance. The Classification task contains 81,788 entries (4.45\%).

The test set (Figure \ref{fig8-task-test}) maintains a parallel distribution across its 226,636 entries. Cross-Modal Identification remains dominant with 175,565 entries (77.47\%), followed by Instance Counting (11,704, 5.16\%), Referring (11,703, 5.16\%), Spatial Grounding (11,608, 5.12\%), Classification (10,024, 4.42\%), and Fine-Grained Description (6,032, 2.66\%). This consistent distribution ensures reliable model evaluation.



The dataset exhibits a well-balanced task distribution that supports comprehensive model training. While the consistency between training and test sets ensures reliable evaluation, two aspects warrant attention: the relatively limited data in Fine-Grained Description tasks may constrain model performance and could benefit from expansion, while the dominant proportion of Cross-Modal Identification data necessitates careful consideration of overfitting during training through appropriate regularization methods.

\subsection{Dataset Examples}

\label{example}

\setlength{\fboxsep}{4pt}  
\setlength{\fboxrule}{0.5pt}  
\noindent\fbox{%
\begin{minipage}{0.46\textwidth}
\small\noindent\textbf{Classification Task:}
\vspace{1pt}
\hrule
\vspace{4pt}

\small\noindent\textbf{User:} {\ttfamily"Please identify the target categories in the figure (candidate categories: \{total category list\})"}

\small\noindent\textbf{AI:} {\ttfamily"The \{specific category\} targets can be identified in the figure"}
\end{minipage}%
}

\setlength{\fboxsep}{4pt}  
\setlength{\fboxrule}{0.5pt}  
\noindent\fbox{%
\begin{minipage}{0.46\textwidth}
\small\noindent\textbf{Spatial Grounding Task:}
\vspace{1pt}
\hrule
\vspace{4pt}

\small\noindent\textbf{User:} {\ttfamily"[grounding] Can you describe the spatial arrangement of various features in this SAR image?"}

\small\noindent\textbf{AI:} {\ttfamily"In this SAR image, I can identify several objects and their locations: There is a \{category-1\} in the top right portion \{bbox\}. There are 2 \{category-2\} in the bottom left: one in the bottom left portion \{bbox\}, another in the middle left portion \{bbox\}. The \{category-1\} is far from the \{category-2\}."}
\end{minipage}%
}

\setlength{\fboxsep}{4pt}  
\setlength{\fboxrule}{0.5pt}  
\noindent\fbox{%
\begin{minipage}{0.46\textwidth}
\small\noindent\textbf{Fine-Grained Description Task:}
\vspace{1pt}
\hrule
\vspace{4pt}

\small\noindent\textbf{User:} {\ttfamily"Please describe the content of the satellite image in detail"}

\small\noindent\textbf{AI:} {\ttfamily"In the figure, \{size\}\{category\} can be seen located at \{orientation\},"+"and there is also \{size\}\{category\} located at \{orientation\}"}
\end{minipage}%
}

\setlength{\fboxsep}{4pt}  
\setlength{\fboxrule}{0.5pt}  
\noindent\fbox{%
\begin{minipage}{0.46\textwidth}
\small\noindent\textbf{Instance Counting Task:}
\vspace{1pt}
\hrule
\vspace{4pt}

\small\noindent\textbf{User:} {\ttfamily"[count] How many \{category\} are there in the image?""}

\small\noindent\textbf{AI:} {\ttfamily"There are 3 instances. \{bbox\}\{bbox\}\{bbox\}"}
\end{minipage}%
}

\setlength{\fboxsep}{4pt}  
\setlength{\fboxrule}{0.5pt}  
\noindent\fbox{%
\begin{minipage}{0.46\textwidth}
\small\noindent\textbf{Cross-Modal Identification Task:}
\vspace{1pt}
\hrule
\vspace{4pt}

\small\noindent\textbf{User:} {\ttfamily"[identify] What object is located in \{bbox\}?"}

\small\noindent\textbf{AI:} {\ttfamily"This is a medium-sized \{category\} located in the top right portion of the image. It appears to be wide in shape."}
\end{minipage}%
}

\setlength{\fboxsep}{4pt}  
\setlength{\fboxrule}{0.5pt}  
\noindent\fbox{%
\begin{minipage}{0.46\textwidth}
\small\noindent\textbf{Referring Task:}
\vspace{1pt}
\hrule
\vspace{4pt}

\small\noindent\textbf{User:} {\ttfamily"[refer] Where is the \{category\}?"}

\small\noindent\textbf{AI:} {\ttfamily"\{bbox\} (top right)."}
\end{minipage}%
}

\section{Motivation for the Dual-benchmark Setup}

In tasks such as spatial localization, cross-modal identification, and referring expression comprehension, this study constructs a dual-benchmark of single-target and multi-target data. The core research motivations are as follows:

\noindent\textbf{(1) Evaluation of the Model's Target Discrimination Ability}

The single-target scenario aims to test the model's basic recognition ability for independent targets. In contrast, the multi-target scenario focuses on examining the model's ability to separate and select targets in complex environments, especially when multiple targets exhibit similar features. This dual-benchmark design can effectively diagnose the performance differences of the model in scenarios with varying degrees of complexity.

\begin{table*}[ht]
    \small
    \setlength{\tabcolsep}{2.5pt}
    \setlength{\arrayrulewidth}{2.5pt} 
    \centering

    \begin{tabular}{lcccc}
        \Thickline
        \textbf{Performance} & \textbf{8B} & \textbf{4B} & \textbf{2B} & \textbf{1B} \\
        \Thickline
        Instance Counting Accuracy & 74.14 & 72.68 & 71.52 & 69.87 \\
        Instance Counting Accuracy (IoU = 0.25) & 61.37 & 57.54 & 54.11 & 50.18 \\
        Instance Counting Accuracy (IoU = 0.5) & 52.17 & 47.35 & 44.22 & 39.35 \\
        Spatial Grounding Accuracy & 62.25 & 60.89 & 60.81 & 56.30 \\
        Abstract Location in Spatial Grounding Accuracy & 81.25 & 83.33 & 50.00 & 0.00 \\
        Spatial Grounding Single Accuracy & 87.91 & 85.90 & 81.92 & 82.24 \\
        Cross-Modal Identification (Multi) Accuracy & 98.84 & 98.01 & 97.79 & 96.98 \\
        Cross-Modal Identification (Single) Accuracy & 98.98 & 98.76 & 98.63 & 98.60 \\
        Referring (Multi) Accuracy (IoU = 0.25) & 37.49 & 34.05 & 27.05 & 22.13 \\
        Referring (Multi) Accuracy (IoU = 0.5) & 23.46 & 18.86 & 13.91 & 9.94 \\
        Referring (Single) Accuracy (IoU = 0.25) & 74.86 & 69.92 & 68.50 & 62.33 \\
        Referring (Single) Accuracy (IoU = 0.5) & 60.13 & 55.29 & 52.16 & 44.99 \\
        Fine-Grained Description Accuracy & 63.43 & 58.84 & 56.36 & 53.30 \\
        Classification Accuracy & 97.25 & 97.27 & 96.69 & 96.65 \\
        \Thickline
    \end{tabular}
        \caption{Performance comparison across different model sizes}
    \label{tab:internvlperformance}
\end{table*}

\noindent \textbf{(2) Revelation of Target Association Understanding Issues}

In multi-target scenarios, the model usually needs to understand the spatial and semantic relationships between targets. By comparing the performance differences of the model in single-target and multi-target scenarios, it is possible to evaluate whether the model truly understands the descriptions of the positional relationships between targets. This helps to identify the limitations of the model when dealing with relative position descriptions such as "the vehicle on the left" and "the tank in the middle".

\noindent\textbf{(3) Exposure of Attention Mechanism Defects}

In multi-target scenarios, the model is highly prone to problems such as attention divergence or overlap. When there are multiple similar targets in an image, the model may have difficulty accurately locating the specific target described by the user. Through the comparison between single-target and multi-target scenarios, the deficiencies of the model in attention allocation can be clearly demonstrated.

\noindent \textbf{(4) Simulation of Real-world Application Scenarios}

Real-world applications cover both simple single-target scenarios and complex multi-target environments. The establishment of the dual-benchmark is more in line with real-world usage requirements, providing a more comprehensive dimension for model evaluation and helping to improve the applicability and reliability of the model in actual deployments. 

\section{Task-specific Performance Scoring}\label{task-cal}

To evaluate model performance on each task, the task-specific accuacy is caculate by Formula \ref{for5}. For each task $t$, the accuracy score $a_{m,t}$ of model $m$ is computed by averaging the accuracy scores across all subtasks:

\begin{equation}
\label{for5}
a_{m,t} = \frac{1}{k} \sum_{i=1}^{k} a_{m,t,i}
\end{equation}

where $a_{m,t}$ is the average accuracy of model $m$ on task $t$, $k$ is the number of subtasks, and $a_{m,t,i}$ is the accuracy of model $m$ on the $i$-th subtask of task $t$. This approach ensures that each subtask contributes equally to the overall task score.
\section{The Analysis of Model Size}
Based on the data analysis in Table \ref{tab:internvlperformance}, it can be concluded that for most task-related metrics, there is a trend of performance improvement as the model size increases from 1B to 8B. For example, the instance-counting accuracy rises from 69.87\% to 74.14\%, the spatial-grounding accuracy increases from 56.30\% to 62.25\%, the fine-grained description accuracy goes up from 53.30\% to 63.43\%, and the classification accuracy climbs from 96.65\% to 97.25\%. This indicates that an increase in model size is beneficial to enhancing the performance of these tasks.
However, the accuracy of abstract locations in the spatial-grounding task shows a unique trend of change. This metric increases from 0.00\% for the 1B model to 83.33\% for the 4B model, but then decreases to 81.25\% for the 8B model, not increasing monotonically with the model size. Evidently, the influence of model size on some specific tasks follows complex patterns. Therefore, when selecting a model, it is necessary to comprehensively consider the task type and model size to achieve optimal performance.

\begin{table*}[htbp]
\scriptsize
\setlength{\tabcolsep}{2.5pt}

\centering
\begin{tabular}{lcccccccccccc}
\Thickline
\multirow{2}{*}{\textbf{Model}} & \multirow{2}{*}{\begin{tabular}[c]{@{}c@{}}\textbf{Only}\\ \textbf{count}\end{tabular}} & \multicolumn{2}{c}{\textbf{Spatial Ground}} & \multicolumn{2}{c}{\textbf{Cross-Modal ID}} & \multicolumn{2}{c}{\textbf{Multi-target Ref}} & \multicolumn{2}{c}{\textbf{Single-target Ref}} & \multirow{2}{*}{\textbf{Descript}} & \multirow{2}{*}{\textbf{Class}} \\
\cline{3-4} \cline{5-6} \cline{7-8} \cline{9-10}
 & & \textbf{Multi} & \textbf{Single} & \textbf{Multi} & \textbf{Single} & \textbf{IoU=.25} & \textbf{IoU=.5} & \textbf{IoU=.25} & \textbf{IoU=.5} & & \\
\Thickline
\textbf{InternVL2.5(8B)} & 63.71 & 5.07 & 0.38 & 0.12 & 0.17 & 0 & 0 & 0 & 0 & 0.04 & 17.8 \\
\textbf{SARChat-InternVL2.5(8B)} & 74.14 & 62.25 & 87.91 & 98.84 & 98.98 & 37.49 & 23.46 & 74.86 & 60.13 & 63.43 & 97.25 \\
\Thickline
\end{tabular}
\caption{Performance comparison of InternVL2.5 before and after fine-tuning}
\label{normal-ft}
\end{table*}

\begin{figure*}[th]
    \centering    \includegraphics[width=0.88\linewidth]{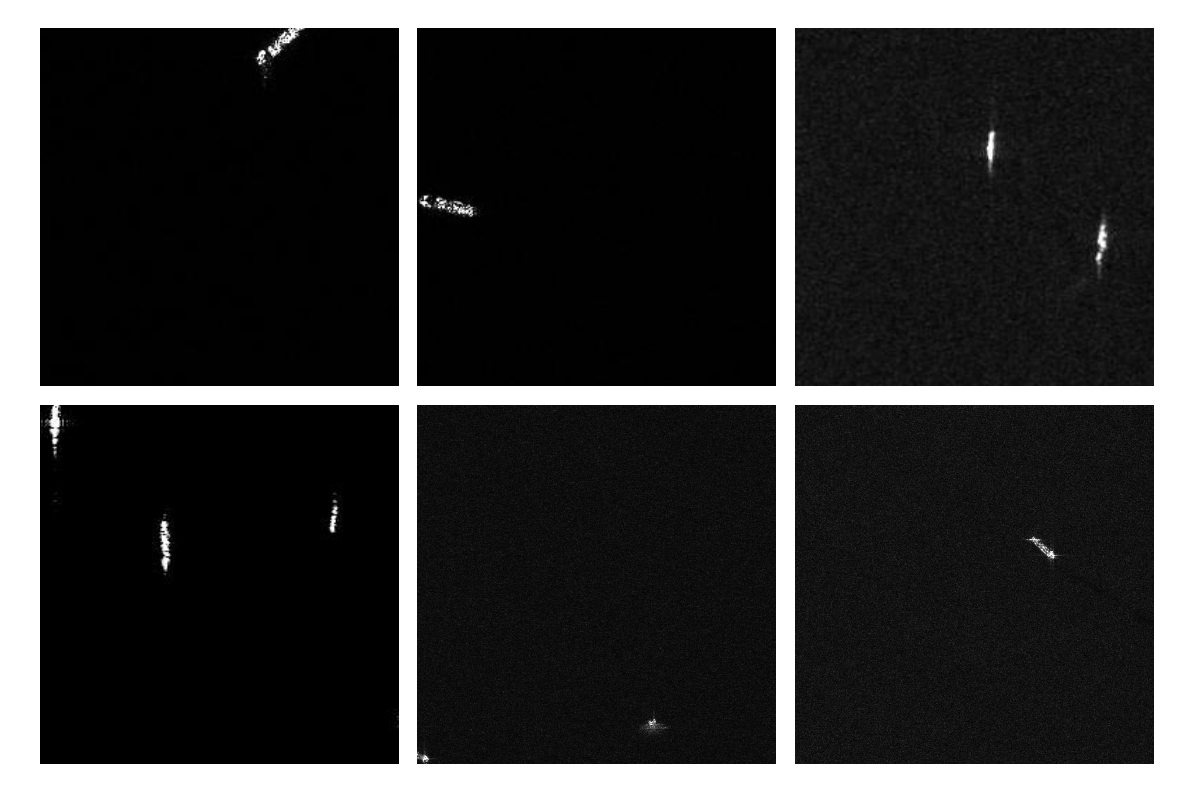}
    \caption{\textbf{Simple Examples of ship detection in SAR images.} The ships appear as distinct bright spots in these SAR images, making them relatively easy to count even for VLMs without SAR-specific training.}
    \label{SAR-sample}
\end{figure*}

\section{Comparison before and after training}
\label{ftornot}
To evaluate the impact of the \ourmethod-2M training dataset, we conducted a comparative analysis on InternVL-2.5-8B—the best-performing model in \ourmethod-Bench—before and after fine-tuning. Our evaluation metrics reveal that without fine-tuning on \ourmethod-2M, the model fails to comprehend most SAR-specific targets, exhibiting near-zero performance on tasks involving target interpretation and description. The only exception is the Instance Counting Task, where InternVL2.5-8B achieves a baseline accuracy of 63.71\%. This relatively high performance can be attributed to the prevalence of ship-on-sea samples, where SAR imaging exhibits distinctive characteristics: the smooth sea surface creates specular reflection, causing most electromagnetic waves to scatter away from the sensor direction, resulting in weak returns that appear as dark areas. Meanwhile, ships' metallic structures and geometric features (such as dihedral and trihedral corner reflectors) generate strong backscattering, concentrating radar waves back to the sensor, thus appearing as bright spots. As shown in Figure \ref{SAR-sample}, these samples present relatively straightforward recognition scenarios, leading to higher counting accuracy scores. The comparative results between the base model and its fine-tuned version on SARChat-Bench are presented in Table \ref{normal-ft}. Overall, fine-tuning with the \ourmethod-2M dataset proves essential for enabling VLMs to interpret SAR imagery effectively.

\end{document}